\documentclass[final]{cvpr}

\usepackage{times}
\usepackage{epsfig}
\usepackage{graphicx}
\usepackage{amsmath}
\usepackage{amssymb}

\usepackage{mmstyle}
\usepackage[utf8]{inputenc} 
\usepackage[T1]{fontenc}    
\usepackage{url}            
\usepackage{booktabs}       
\usepackage{amsfonts}       
\usepackage{nicefrac}       
\usepackage{microtype}      
\usepackage{color}

\usepackage[position=top]{subfig}

\usepackage[pagebackref=true,breaklinks=true,colorlinks,bookmarks=false]{hyperref}

\begin{document}

\title{Video Representation Learning with Visual Tempo Consistency}
\author{
   Ceyuan Yang$^1$, Yinghao Xu$^1$, Bo Dai$^2$, Bolei Zhou$^1$\\
  $^1$The Chinese University of Hong Kong, $^2$Nanyang Technological University\\
  {\tt\small \{yc019, xy119, bzhou\}@ie.cuhk.edu.hk},
  {\tt\small bo.dai@ntu.edu.sg}\\
}

\maketitle

\begin{abstract}
   Visual tempo, which describes how fast an action goes, has shown its potential in supervised action recognition \cite{slowfast,tpn}.
In this work, we demonstrate that visual tempo can also serve as a self-supervision signal for video representation learning.
We propose to maximize the mutual information between representations of \emph{slow} and \emph{fast} videos via hierarchical contrastive learning (VTHCL).
Specifically, by sampling the same instance at slow and fast frame rates respectively, we can obtain slow and fast video frames which share the same semantics but contain different visual tempos.
Video representations learned from VTHCL achieve the competitive performances under the self-supervision evaluation protocol for action recognition on UCF-101 (82.1\%) and HMDB-51 (49.2\%).
Moreover, comprehensive experiments suggest that the learned representations are generalized well to other downstream tasks including action detection on AVA and action anticipation on Epic-Kitchen.
Finally,
we use Instance Correspondence Map (ICM) to visualize the shared semantics captured by contrastive learning.\footnote{Code and models are available at \href{https://github.com/decisionforce/VTHCL}{this link.}}

   \end{abstract}
   
   \section{Introduction}\label{sec:intro}
In recent years, a great success of representation learning has been made, especially for self-supervised learning from images.
The visual features obtained in a self-supervised manner have been getting very close to those of supervised training on ImageNet \cite{iamgenet}.
Meanwhile, representing videos in a compact and informative way is also crucial for many analysis,
since videos are redundant and noisy in their raw forms.
However, supervised video representation learning demands a huge number of annotations,
which in turn encourages researchers to investigate self-supervised learning schemes to harvest the massive amount of unlabelled videos.

\begin{figure}[t]
	\centering
	\includegraphics[width=1.0\linewidth]{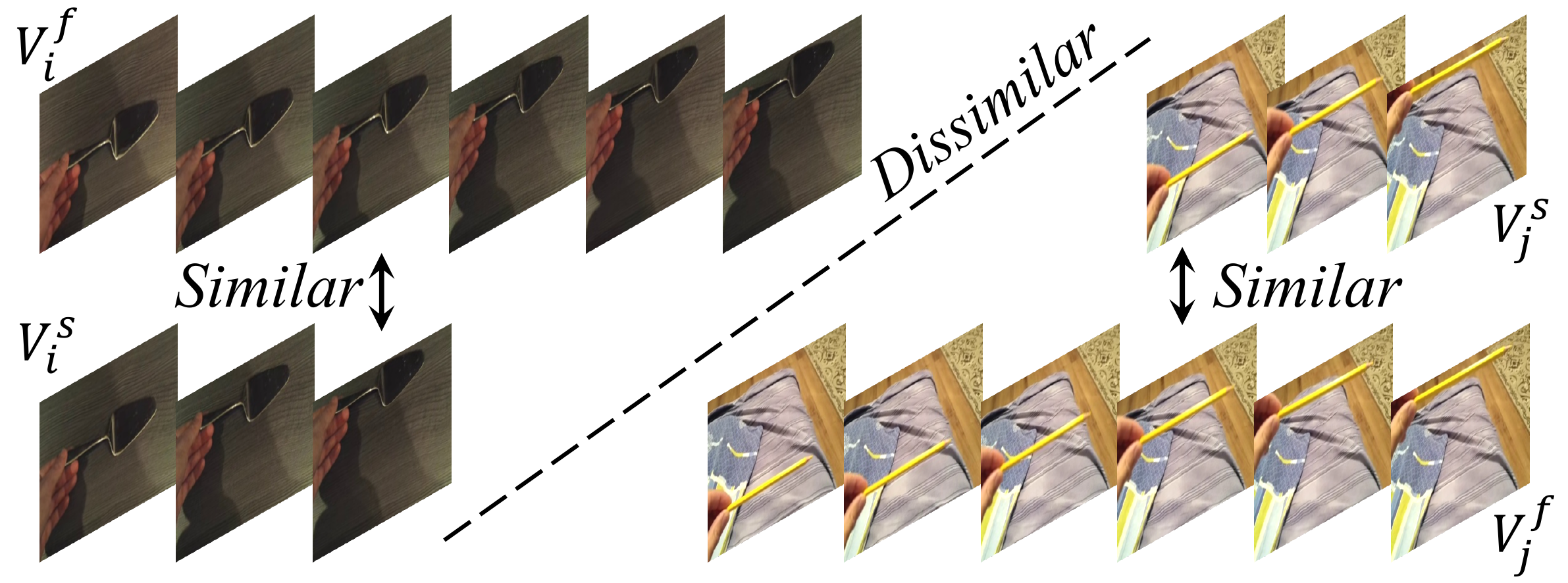}
	\captionsetup{font=small}
	\caption{
       \textbf{Visual Tempo Consistency} enforces the network to learn the high representational similarity between the same instance (\eg $V_i$) sampled at different tempos (\eg $V_{i}^s$ and $V_{i}^f$). Meanwhile, it also follows the same mechanism as previous instance discrimination task \cite{wu2018unsupervised} which distinguishes individual instances according to the visual cues
	}
    \label{fig:teaser}
  \end{figure}

Videos contain rich motion dynamics along the temporal dimension.
Thus, if we can make the best of the underlying consistency as well as causality dependency in the activities occurring in the videos, we can better leverage a large amount of unlabelled data for representation learning.
For instance, previous attempts learn video representations by predicting the correct order of shuffled frames \cite{shuffle_and_learn},
the arrow of time \cite{arrow_of_time}, and the frames and motion dynamics in the future \cite{predict_future_motion,dpc}.
Considering the recent success of exploiting visual tempo in action recognition tasks \cite{slowfast,tpn},
in this work we aim at exploring visual tempo for self-supervised video representation learning.

Visual tempo, which describes how fast an action goes, is an essential variation factor of video semantics.
Particularly, an action instance can be performed and observed at different tempos due to multiple elements
including mood and age of the performer and configuration of the observer, the resulting video thus varies from case by case.
Nonetheless, the same instance with different tempos is supposed to share high similarity
in terms of their discriminative semantics, which is exactly the underlying consistency for self-supervised representation learning.

While visual tempo could be utilized by directly predicting the correct tempo of a given action instance
as in previous attempts \cite{speediness,shuffle_and_learn,arrow_of_time,predict_future_motion},
we argue that such a predictive approach may enforce the learned representations to capture the information that distinguishes the frequency of visual tempos,
which is not necessarily related to the discriminative semantics we are looking for.
Therefore, we propose an alternative approach based on contrastive learning \cite{hadsell2006dimensionality,he2019momentum,cmc,mocov2},
which maximizes the mutual information between representations across videos from the same action instance but with different visual tempos.
Specifically, we formulate self-supervised video representation learning
as the consistency measurement between a pair of videos,
which contains video frames from the same action instance but being sampled at the \emph{slow} and \emph{fast} visual tempo respectively.
As is shown in Fig. \ref{fig:teaser}, the learning is conducted by adopting a \emph{slow} and a \emph{fast} video encoder,
and taking in turn a video from each pair as the query to distinguish its counterpart from a set of negative samples.
In this way, the resulting video representations are expected to capture the shared information and better retain its discriminations.
Additionally, considering the key of such learning is the shared information, we develop Instance Correspondence Map (ICM) to visualize the shared information captured by contrast learning.

As shown in the literature \cite{tpn} that the feature hierarchy inside a video network (\eg~I3D \cite{k400})
already reflects semantics at various visual tempos, we further propose a hierarchical contrastive learning scheme,
where we use network features across multiple depths as queries.
Such a scheme not only leverages the variation of visual tempo more effectively but also provides a stronger supervision for deeper networks.
Evaluated thoroughly on a wide variety of downstream action understanding tasks including action recognition on UCF-101 \cite{ucf101} and HMDB-51 \cite{hmdb},
action detection on AVA \cite{ava}, and action anticipation on Epic-Kitchen \cite{epic-kitchen},
we find the representations learned via exploiting visual tempo consistency are highly discriminative and generalizable,
leading to the competitive performances. 

We summarize our contributions as follows:
a) We demonstrate visual tempo can serve as a strong supervision signal for unsupervised video representation learning, which is utilized by the proposed hierarchical contrastive learning scheme.
b) We show that our proposed framework can achieve competitive performances for action recognition on UCF-101 and HMDB-51, and generalize well to other downstream tasks such as action detection and action anticipation.
c) We propose Instance Correspondence Map to qualitatively interpret the learned representations, which highlights the informative objects in videos.

   \section{Related Work}\label{related}
\noindent \textbf{Self-supervised Video Representation Learning.}
Various pretext tasks have been explored for self-supervised video representation learning,
such as modeling the cycle-consistency between two videos of the same category \cite{dwibedi2019temporal},
modeling the cycle-consistency of time \cite{wang2019learning},
predicting the temporal order of frames \cite{odd-one-out,sort-seq,shuffle_and_learn,arrow_of_time},
predicting future motion dynamics and frames \cite{predict_future_motion,dpc,oord2018representation}
as well as predicting the color of frames \cite{vondrick2018tracking}.
In this work, we explore a different pretext task,
which models the consistency between videos from the same action instance but with different visual tempos.
There are works that learn video representations using not only videos themselves
but also corresponding text \cite{sun2019contrastive,sun2019videobert,miech2019end}
and audios \cite{korbar2018cooperative,arandjelovic2017look,alwassel2019self,patrick2020multi}.
Besides, cocurrent work \cite{cotraining} proposed a co-training scheme to learn representations from RGB and Optical Flow.
In contrast to those, we learn compact representations from RGB frames only.

\noindent \textbf{Contrastive Learning.}
Due to their promising performances,
contrastive learning and its variants \cite{bachman2019learning,henaff2019data,hjelm2018learning,oord2018representation,cmc,wu2018unsupervised,he2019momentum,chen2020simple} are considered as an important direction for self-supervised representation learning.
Particularly, the most related work is the contrastive multiview coding \cite{cmc},
which learns video representations by maximizing the mutual information between RGB and flow data of the same frames.
The difference is that in this work we learn video representations via the consistency between videos of the same action instance but with different visual tempos.
Moreover, we further introduce a hierarchical scheme to leverage such consistency at different depths of the encoding network,
providing a stronger supervision for training deeper networks.

\noindent \textbf{Representation Interpretation.}
Interpreting what the deep neural networks have learned gives insight into the generalization ability and the transferability of deep features \cite{morcos2018importance, yosinski2014transferable}.
Particularly, some of them \cite{zhou2014object, bau2017network} developed techniques to study the hidden units.
Besides, mapping a given representation back to image space \cite{simonyan2013deep, zeiler2014visualizing,cam, nguyen2016synthesizing,mahendran2015understanding} also explain what CNNs actually learn to distinguish different categories.
However, these techniques cannot be directly applied to representations learned from contrastive learning since there are no semantic categories during training.
In this work, without regarding the categorical annotations, we develop Instance Correspondence Map to qualitatively interpret the correspondences at the instance-level as a way to reveal the shared information learned by our method.

   \section{Learning from Visual Tempo Consistency}\label{sec:method}
The goal of self-supervised video representation learning is to learn a video encoder $g$ that is able to produce compact and informative video representations,
by regarding the structural knowledge and the consistency among a set of unlabelled videos $\{v_1, ..., v_n\}$ as the self-supervision signal.
The discriminative feature of $g$ is often verified through a set of downstream tasks (\eg action classification, action detection and action anticipation).
While various supervisions have been proposed by previous attempts, in this work we introduce the visual tempo consistency,
a novel and effective self-supervision signal. 
We start by discussing what is the visual tempo consistency and why it is a strong supervision signal, then we introduce its learning process.

\begin{figure*}[t]
	\centering
	\includegraphics[width=1.0\textwidth]{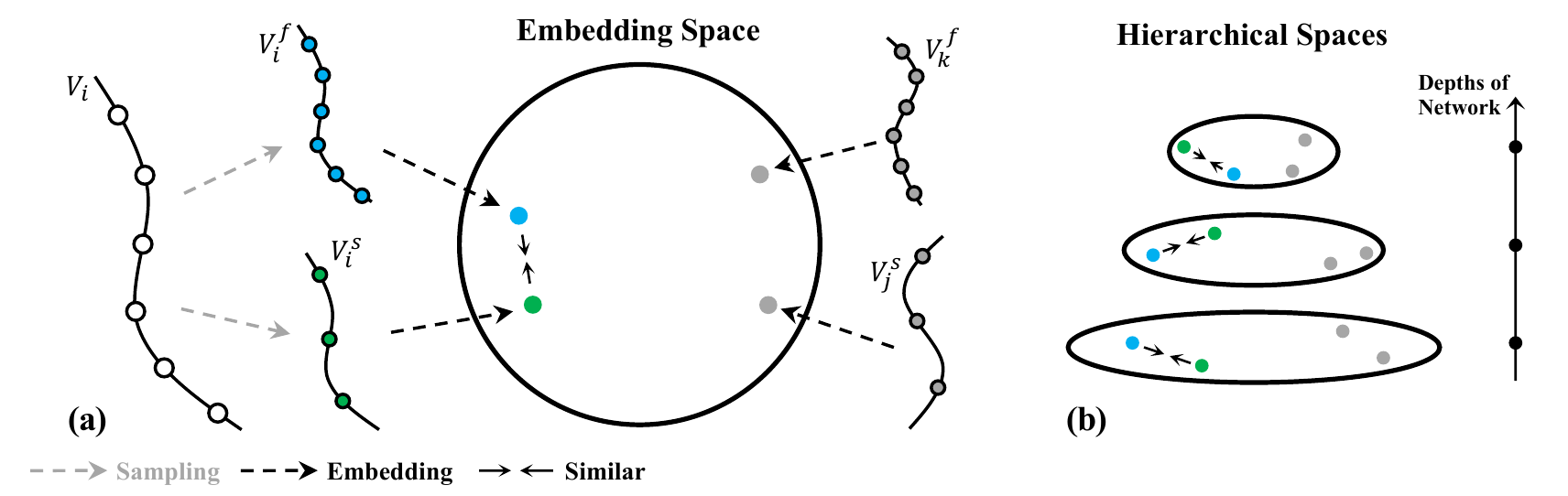}
	\captionsetup{font=small}
	\caption{
	   \textbf{Framework.}
	   (a) The same instance with different tempos (\eg $V_{i}^f$ and $V_{i}^s$) should share high similarity in terms of their discriminative semantics while are dissimilar to other instances (grey dots).
	   (b) The features at various depths of networks allow to construct the hierarchical representation spaces
	}
    \label{fig:pipeline}
  \end{figure*}

\subsection{Visual Tempo Consistency as a Self-supervision Signal}\label{subsec:visual_tempo}
Following \cite{tpn}, we refer to visual tempo as how fast an action goes in an action video.
As an internal variation factor of these videos, the visual tempos of actions across different classes have a large variance.
In previous literature \cite{tpn,slowfast}, the benefits of considering the variance of visual tempo in supervised recognition tasks have been well explored.
A question then arises: \emph{Whether such variance can also benefit self-supervised learning?}
With a proper formulation of the variance of visual tempo, we show that it can serve as an effective and promising self-supervision signal.

Specifically, as shown in \cite{slowfast} we can adjust the sampling rate of frames to get videos of the \emph{same} action instance but with \emph{different} visual tempos.
Without the loss of generality, we use videos of two different sampling rates and refer to them as fast and slow videos,
\ie~$\cV_f = \{v_{1,f}, ..., v_{n,f}\}$ and $\cV_s = \{v_{1,s}, ..., v_{n,s}\}$.
In order to ignore the effect of different distribution on backbones \cite{huang2018makes},
we thus introduce two encoders, $g_f$ and $g_s$, respectively for fast and slow videos
and learn them by matching the representations of an action instance's corresponding fast and slow videos.

The intuition behind such approach for video representation learning is that, at first, learning via the consistency between multiple representations is shown to be more effective than learning by prediction \cite{cmc,he2019momentum,mocov2,chen2020simple}.
Moreover, while previous attempts resort to matching representations of different patches \cite{wu2018unsupervised} or different views (\eg~RGB and optical flow) \cite{cmc} of the same instance,
the inputs of these representations intrinsically have different semantics.
On the contrary, the semantics of an instance's fast and slow videos are almost identical, with visual tempo being the only difference.
Encouraging the representation consistency between videos of the same instance but with different visual tempos thus provides a stronger supervision signal.

\subsection{Adopting Visual Tempo Consistency via Contrastive Learning}\label{subsec:contrastive}

We apply contrastive learning to train our encoders $g_f$ and $g_s$.
Specifically, given two sets of videos $\cV_f = \{v_{1, f}, ..., v_{n, f}\}$ and $\cV_s = \{v_{1, s}, ..., v_{n, s}\}$,
where $i$-th pair of videos $(v_{i, f}, v_{i, s})$ contains two videos of the same $i$-th instance but with different visual tempos,
we can get their corresponding representations $\cX_f = \{\vx_{1,f}, ..., \vx_{n,f}\}$ and $\cX_s=\{\vx_{1,s}, ..., \vx_{n,s}\}$ by
\begin{align}\label{eqa:extract_feature}
	\vx_{i, f} & = g_f(v_{i, f}), \\
	\vx_{i, s} & = g_s(v_{i, s}),
\end{align}
where we refer to $\vx_{i, f}$ and $\vx_{i, s}$ as the fast and slow representations of $i$-th instance.
Learning $g_f$ and $g_s$ based on the visual tempo consistency involves two directions.
For each fast representation $\vx_{i, f}$, we encourage the similarity between $\vx_{i, f}$ and its slow representation counterpart $\vx_{i, s}$ while decreasing the similarities between it and other slow representations.
This process also applies to each slow representation.
Subsequently we can obtain the loss functions:
\begin{align}\label{eqa:cl_loss}
	\cL_f & = - \sum_{i=1}^n \left[ \log \frac{ \exp (h(\vx_{i, f}, \vx_{i, s}))}{\sum_{j=1}^n \exp(h(\vx_{i, f}, \vx_{j, s}))} \right], \\
	\cL_s & = - \sum_{i=1}^n \left[ \log \frac{ \exp (h(\vx_{i, f}, \vx_{i, s}))}{\sum_{j=1}^n \exp(h(\vx_{j, f}, \vx_{i, s}))} \right], \\
	\cL_\mathrm{total} & = \cL_f + \cL_s,
\end{align}
where $h$ is a function measuring the similarity between two representations.
$h$ can be calculated by
\begin{align}\label{eqa:sim}
h(\vx_{i, f}, \vx_{i, s}) = \frac{\phi(\vx_{i, f}) \cdot \phi(\vx_{i, s})}{T \cdot \Vert \phi(\vx_{i, f}) \Vert_2 \cdot \Vert \phi(\vx_{i, s}) \Vert_2}.
\end{align}
Here $T$ is the temperature hyperparameter~\cite{wu2018unsupervised}, and $\phi$ is a learnable mapping.
As suggested by \cite{chen2020simple, mocov2}, applying a non-linear mapping function can substantially improve the learned representations.

\noindent \textbf{Memory bank.} It is non-trivial to scale up if we extract the features of all videos at each iteration.
Consequently, we reduce the computation overhead by maintaining two memory banks $\cB_f$ and $\cB_s$ of size $n \times d$ as in \cite{wu2018unsupervised,cmc} where $d$ is the dimension of representations.
$\cB_f$ and $\cB_s$ respectively store the approximated representations of fast and slow videos.
Representations stored in $\cB_f$ and $\cB_s$ are accumulated over iterations as
\begin{align}
\vx_\mathrm{bank} = m \vx_\mathrm{bank} + (1 - m) \vx_\mathrm{current},
\end{align}
where $\vx$ can be any $\vx_{i,f}$ or $\vx_{i,s}$, and $m \in [0, 1]$ is the momentum coefficient to ensure smoothness and stability.
Based on $\cB_f$ and $\cB_s$, the learning process thus becomes taking a mini-batch of fast video as queries, computing the loss function $\cL_f$ based on their representation obtained via $g_f$ and $N$ sampled representations stored in $\cB_s$.
$\cL_s$ can be computed in a similar manner.
It is worth noting one can further reduce the computation overhead by sampling $m$ representations from each bank rather than using the entire bank when computing $\cL_f$ and $\cL_s$,
or adopting noise contrastive estimation as in \cite{wu2018unsupervised, cmc}.

\subsection{Learning from Visual Tempo via Hierarchical Contrastive Learning}\label{subsec:hierarchical_contrastive}
While we usually use the final output of $g_f$ and $g_s$ as the representation of an input video,
it is known \cite{slowfast,tpn} that popular choices of $g_f$ and $g_s$ (\eg I3D \cite{k400})
contain a rich temporal hierarchy inside their architectures, \ie~features of these networks at different depths already encode various temporal information due to their varying temporal receptive fields.
Inspired by this observation, we propose to extend the loss functions in Eq.\eqref{eqa:cl_loss} to a hierarchical scheme,
so that we can provide $g_f$ and $g_s$ a stronger supervision.
The framework is shown in Fig.\ref{fig:pipeline}. Particularly, the original contrastive learning can be regarded as a special case where only the final feature is used.
Specifically, we use features at different depths of $g_f$ and $g_s$ as multiple representations of an input video,
\ie~replacing $\vx_{i, f}$ and $\vx_{i, s}$ of $i$-th fast and slow videos with $\{\vx_{i, f}^k\}_{k \in \cK}$ and $\{\vx_{i, s}^k\}_{k \in \cK}$,
where $\cK$ is the set of depths we choose to extract features from $g_f$ and $g_s$.
For instance, we could collect the output of each residual layers (\ie $\{res_i\}_{i=2}^5$) in 3D-ResNet \cite{k400} to construct set $\cK$.
Accordingly, the original two memory banks are extended to a total of $2|\cK|$ memory banks,
and the final loss function is extended to $\cL_\mathrm{total} = \sum_{k \in \cK} \lambda^k (\cL^k_f + \cL^k_s)$,
where
\begin{align}\label{eqa:cl_loss}
	\forall k \in \cK \qquad
	\cL^k_f & = - \sum_{i=1}^n \left[ \log \frac{ \exp (h(\vx^k_{i, f}, \vx^k_{i, s}))}{\sum_{j=1}^n \exp(h(\vx^k_{i, f}, \vx^k_{j, s}))} \right], \\
	\cL^k_s & = - \sum_{i=1}^n \left[ \log \frac{ \exp (h(\vx^k_{i, f}, \vx^k_{i, s}))}{\sum_{j=1}^n \exp(h(\vx^k_{j, f}, \vx^k_{i, s}))} \right].
\end{align}

\section{Instance Correspondence Map}\label{sec:icm}

\begin{figure}[t]
	\centering
	\includegraphics[width=1.0\linewidth]{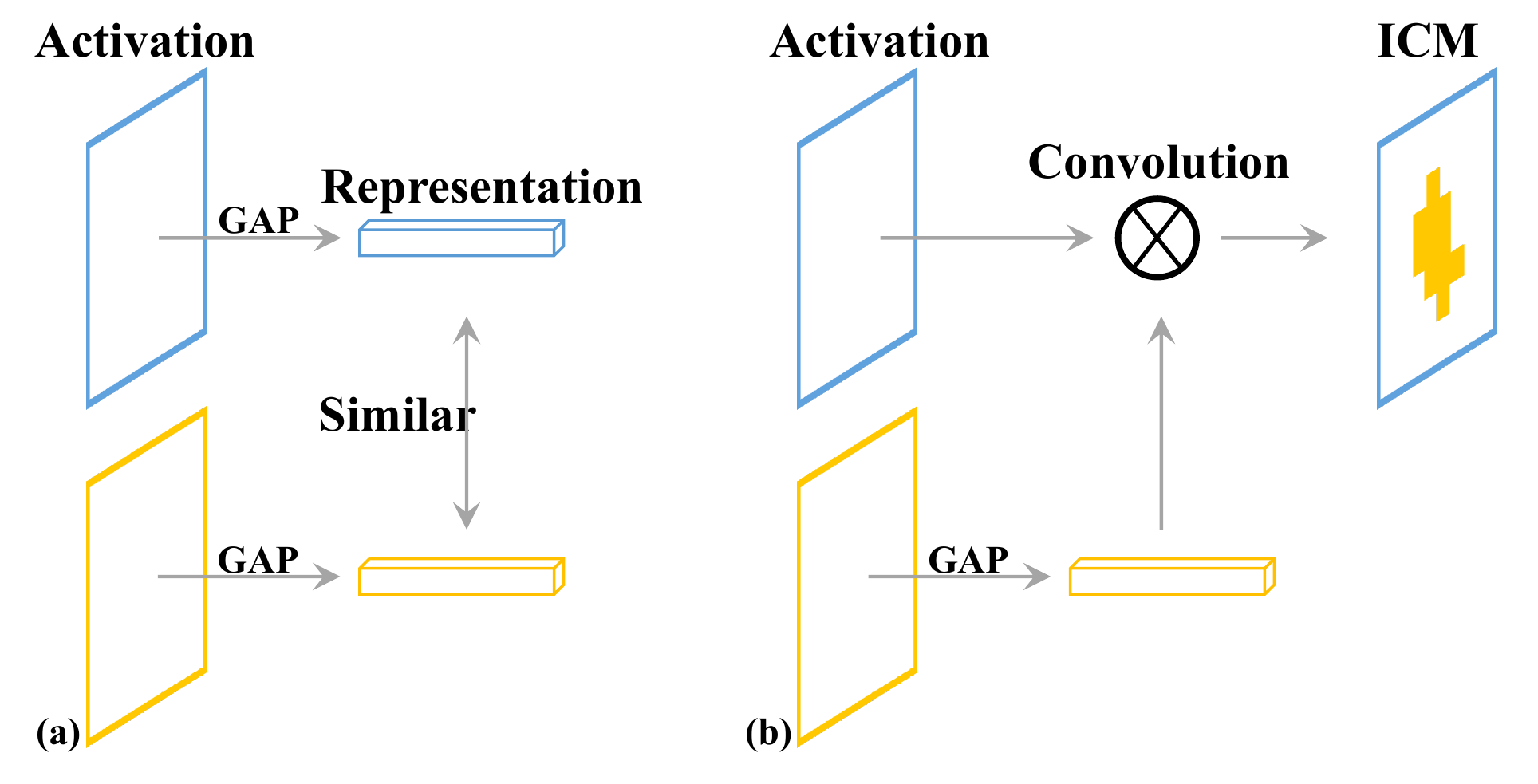}
	\captionsetup{font=small}
	\caption{
	   \textbf{Illustration of ICM.}
	   (a) The similarity measurement between a positive pair (the blue and yellow).
	   (b) Using one sample in the pair as the reference, ICM highlights the instance-specific shared regions. Note that the channel and temporal dimensions are omitted for brevity
	}
    \label{fig:cam}
  \end{figure}

Recall the formulation in Sec.\ref{subsec:contrastive}, a pair of videos $(v_{i, f}, v_{i, s})$ are videos of the same instance but with different visual tempos.
The corresponding representations could be obtained by two encoders $g_f$ and $g_s$.
Particularly, the last operation of encoders is global-avg-pooling (GAP) which averages the activation of each element on feature maps. Namely, the representations $\vx_{i, f}$ and $\vx_{i, s}$ are derived as
\begin{align}\label{eqa:avg}
	\vx_{i, f} & = GAP(x_{i, f}), \\
	\vx_{i, s} & = GAP(x_{i, s}),
\end{align}
where $x_{i, f}$ and $x_{i, s}$ are the activations before the last operation of encoders (\ie GAP) with the shape of $C \times T_f \times W \times H$ and $C \times T_s \times W \times H$ respectively.
We could regard one of the two final representations as the instance-aware classifier to calculate the pointwise similarity of the other activation map.
Therefore, the instance correspondence map ($M_i$) could be calculated via the pointwise multiplication between the activation map ($x_{i,s}$) and the final representation ($\vx_{i,f}$).
Figure \ref{fig:cam} presents how the Instance Correspondence Map (ICM) is calculated.
Technically, final representations with spatial and temporal dimensions could be easily obtained by converting the fc layer in the learnable mapping $\phi$ to a convolutional layer with kernel size $1\times 1$.
%

   \section{Experiments}\label{sec:exp}
We conduct a series of comprehensive experiments following the standard protocol of evaluating video representations from self-supervised learning.
Specifically, we pretrain video encoders with the proposed VTHCL on a large-scale dataset (\eg~Kinetics-400 \cite{k400}) then finetune the encoders on the target dataset corresponding to a certain downstream task (\eg~UCF-101 and HMDB-51 for action recognition).
In practice, we regard the encoder $g_s$ for slow videos as our main encoder used for evaluation.
To ensure reproducibility, all implementation details are included in Sec.\ref{subsec:details}.
Main results of action recognition are presented in Sec.\ref{subsec:main_results} with comparison to prior approaches.
Sec.\ref{subsec:ablation} includes ablation studies on the components of VTHCL.
To further demonstrate the effectiveness of VTHCL and show the limitation of current evaluation protocol,\
we evaluate VTHCL on a diverse downstream tasks including action detection on AVA \cite{ava} and action anticipation on Epic-Kitchen \cite{epic-kitchen} in Sec.\ref{subsubsec:ava} and Sec.\ref{subsubsec:epic} respectively.
Finally, we also interpret the learned representations via ICM in Sec.\ref{subsec:interpretation}.
It is worth noting all experiments are conducted on a single modality (\ie~RGB frames) and evaluated on the corresponding validation set unless state otherwise.

\subsection{Implementation Details}\label{subsec:details}

\noindent \textbf{Backbone.}
Two paths of SlowFast \cite{slowfast} without lateral connections are adapted as $g_f$ and $g_s$,
which are modified from 2D ResNet \cite{resnet} by inflating 2D kernels \cite{k400}.
The main difference between two encoders is the network width and the number of inflated blocks.
Importantly, after training, only the slow encoder would be adopted for various tasks.

\begin{table*}[t]
    \caption{
        \textbf{Comparison with other state-of-the-art methods on UCF-101 and HMDB-51.} Note that only the top-1 accuracies are reported 
    }
    \label{table:sota}
    \centering
    \setlength{\tabcolsep}{4.0mm}{
    \begin{tabular}{llccc}
      \toprule
      Method                             & Architecture  & \#Frames  & UCF-101 \cite{ucf101} & HMDB-51 \cite{hmdb} \\
      \midrule
      Random/ImageNet/Kinetics           &  3D-ResNet18  & 8        & 68.0/83.0/92.6   & 30.8/48.2/66.7\\
      Random/ImageNet/Kinetics           &  3D-ResNet50  & 8        & 61.1/86.2/94.8   & 21.7/51.8/69.3\\
      \midrule
      MotionPred \cite{motionpred}       &  C3D          & 16       & 61.2 & 33.4 \\
      RotNet3D \cite{rotnet3d}           &  3D-ResNet18  & 16       & 62.9 & 33.7 \\
      ST-Puzzle \cite{stpuzzle}          &  3D-ResNet18  & 16       & 65.8 & 33.7 \\
      ClipOrder \cite{cliporder}         &  R(2+1)D-18   & -        & 72.4 & 30.9 \\
      DPC \cite{dpc}                     &  3D-ResNet34  & -        & 75.7 & 35.7 \\
      AoT \cite{arrow_of_time}           &  T-CAM        & -        & 79.4 & -    \\
      SpeedNet \cite{speediness}         &  I3D          & 64       & 66.7 & 43.7 \\
      PacePrediction \cite{wang2020self} &  R(2+1)D-18   & -        & 77.1 & 36.6 \\
      \midrule
      VTHCL-R18 (Ours)                   &  3D-ResNet18  & 8        & 80.6 & 48.6 \\
      VTHCL-R50 (Ours)                   &  3D-ResNet50  & 8        & \textbf{82.1} & \textbf{49.2} \\
      \bottomrule
    \end{tabular}}
  \end{table*}

\noindent \textbf{Training Protocol.}
Following \cite{cmc,he2019momentum,wu2018unsupervised,chen2020simple}, video encoders in VTHCL are randomly initialized as default.
Synchronized SGD serves as our optimizer, whose weight decay and momentum are set to 0.0001 and 0.9 respectively.
The initial learning rate is set to 0.03 with a total batch size of 256.
The half-period cosine schedule \cite{cosine} is adapted to adjust the learning rate (200 epochs in total).
Following the hyperparameters in \cite{wu2018unsupervised,cmc}, temperature $T$ in Eq.\eqref{eqa:sim} is set to 0.07 and the number of sampled representation $N$ is 16384.

\noindent \textbf{Dataset.} Kinetics-400 \cite{k400} contains around 240k training videos which serve as the large-scale benchmark for self-supervised representation learning.
We extract video frames at the raw frame per second (FPS) and sample the consecutive 64 frames as a raw clip which can be re-sampled to produce slow and fast clips at the specific stride $\tau$ and $\tau / \alpha$ ($\alpha > 1$) separately.
Unless state otherwise, the sample stride $\tau$ is 8, \ie~our model will take 8 frames ($8=64/8$) as input.

\subsection{Action Recognition}\label{subsec:main_results}
\noindent \textbf{Setup.}
In order to conduct a fair comparison, following prior works we finetune the learned video encoders of VTHCL on UCF-101 \cite{ucf101} and HMDB-51 \cite{hmdb} datasets for action recognition.
Particularly, we obtain the video accuracy via the standard protocol \cite{slowfast,tpn,nonlocal}, \ie~uniformly sampling 10 clips of the whole video and averaging the softmax probabilities of all clips as the final prediction.
We train our models for 100 epochs with a total batch size of 64 and an initial learning rate of 0.1, which is reduced by a factor of 10 at 40, 80 epoch respectively.
Moreover, when pre-training on Kinetics-400 \cite{k400}, three levels of contrastive hierarchy is constructed, \ie~we collect features from the output of $\{res_3, res_4, res_5\}$ due to the limitation of GPU resources.
Unless state otherwise, $\alpha$ is defaultly set to 2 for the fast clips (sample stride of fast encoder $g_f$ is $8/2=4$).
Namely, the slow and fast encoders take 8 and 16 frames as the input separately.

\noindent \textbf{Main Results.}
Table \ref{table:sota} illustrates the comparison between ours and other previous approaches. Here all the methods utilize only a single modality and similar architectures.
Besides, the results using different types of initializations (\ie~Random, ImageNet inflated and Kinetics pretrained) are also included to serve as the lower/upper bounds.
In particular, our method equipped with the shallower network (3D-ResNet18) can achieve top-1 accuracy of 80.6\% and 48.6\% respectively, outperforming previous works with a similar setting by large margins.
Furthermore, increasing the capacity of the network from 3D-ResNet18 to 3D-ResNet50 can introduce a consistent improvement, achieving 82.1\% and 49.2\% top-1 accuracies.
Compared to the supervised results of similar backbones obtained using a random initialization (\eg~61.1\% and 68.0\% on UCF-101 \cite{ucf101} for 3D-ResNet18 and 3D-ResNet50), our method can significantly decrease the gap between self-supervised and supervised video representation learning.

\noindent \textbf{Effect of Architectures.}
Beyond the competitive performances, Table \ref{table:sota} also raises the awareness of the effect of various backbones.
Intuitively, when increasing network capacity, the learned representations should be better.
For example, works in image representation learning \cite{revisitssl,he2019momentum,mocov2,cmc} confirms networks with larger capacities can boost the quality of learned representations.
As for video representation learning, it can be seen from Table \ref{table:sota}, when networks are well initialized (\eg~supervised pretraining on ImageNet and Kinetics, or using VTHCL on Kinetics), the one with a larger capacity indeed outperforms its counterpart.
Particularly, when randomly initialized, 3D-ResNet50 performs worse on UCF-101 and HMDB than 3D-ResNet18 although it has a relatively larger capacity.
It indicates the number of parameters of 3D-ResNet50 is too large compared to the scale of UCF-101 and HMDB, so that it suffers from overfitting.
Therefore, while prior works usually employed a relatively shallow model (\eg~3D-ResNet18) in the evaluation, it is important to test a heavy backbone to see whether the proposed methods perform consistently across backbones.

\begin{table}[t]\centering
    \captionsetup[subfloat]{captionskip=2pt}
    \caption{
        \textbf{Ablation Studies on visual tempo and hierarchical contrastive formulation.} We report the top-1 accuracy on UCF-101 \cite{ucf101} and HMDB-51 \cite{hmdb} respectively
    }
    \subfloat[\textbf{Various visual tempo.} $\alpha$ denotes the relative coefficient of sample stride for fast clip\label{table:ablation:visual_tempo}]
        {
        \begin{tabular}{lccc}
            \toprule
            Models        &   $\alpha=1$  &$\alpha=2$ & $\alpha=4$            \\
            \midrule
            R18   &  78.2/45.2    & 79.5/47.4 & 80.0/48.2           \\
            R50   &  80.3/47.3    & 80.9/47.7 & 80.6/48.0            \\
            \bottomrule
        \end{tabular}
        }\hspace{0.5mm}
    \subfloat[\textbf{Various levels of contrastive formulation.} $D$ denotes the number of levels of contrastive formulation\label{table:ablation:hcl}]
        {
        \begin{tabular}{lccc}
            \toprule
            Models        &   $D=1$  &$D=2$ & $D=3$            \\
            \midrule
            R18   &  79.5/47.4    & 80.3/47.9 & 80.6/48.6            \\
            R50   &  80.9/47.7    & 81.5/48.5 & 82.1/49.2             \\
            \bottomrule
        \end{tabular}
        }
    \vspace{-1em}

    \label{table:ablations}
\end{table}

\subsection{Ablation Studies}\label{subsec:ablation}
Here we include the ablation study to investigate the effect of different VTHCL components.

\noindent \textbf{Effect of relative visual tempo difference.}
Although in Table \ref{table:sota} we show VTHCL can obtain competitive results on UCF-101 \cite{ucf101} and HMDB \cite{hmdb}, it remains uncertain whether the relative visual tempo difference between slow and fast videos significantly affects the performance of VTHCL.
We thus conduct multiple experiments by adjusting the relative coefficient of sample stride (\ie $\alpha = \{1,2,4\}$).
Specifically, 8, 16 and 32 frames are respectively fed into fast encoder $g_f$ while maintaining the number of frames for slow encoder $g_s$ as 8.
When $\alpha$ is 1, the input is exactly the same for both slow and fast encoders.
In this case, VTHCL actually turns into instance discrimination task \cite{wu2018unsupervised} which distinguishes video instances mainly via the appearance instead of utilizing visual tempo consistency.
Such a setting thus serves as our baseline to tell whether the visual tempo could help learn better video representations.
Moreover, to avoid unexpected effects, we do not apply the hierarchical scheme, and only the final features of two encoders are used as in Sec.\ref{subsec:contrastive}.

Results are included in Table.\ref{table:ablation:visual_tempo}, which suggests that a larger $\alpha$ generally leads to a better performance for both 3D-ResNet18 and 3D-ResNet50.
It has verified that the visual tempo difference between slow and fast videos indeed enforces video encoders to learn discriminative semantics utilizing the underlying consistency.
Visual tempo as a source of the supervision signal can help self-supervised video representation learning.

\noindent \textbf{Effect of hierarchical contrastive learning.}
We study the effect of the hierarchical contrastive formulation with a varying number of levels.
Here $D$ refers to the number of elements in $\cK$. For example, we collect the features from $\{res_4, res_5\}$ and build up a two-level contrastive formulation when $D=2$.
Furthermore, when $D$ is 1, the hierarchical scheme degrades into the general contrastive formulation shown in Sec.\ref{subsec:contrastive}.
The relative coefficient $\alpha$ is set to 2 for a fair comparison.

Results are included in Table.\ref{table:ablation:hcl}, showing that an increasing number of levels in the contrastive formulation significantly boosts the performance even when the model is quite heavy and tends to overfit.
These results verify the effectiveness of utilizing the rich hierarchy inside a deep network, which correlate well with previous studies \cite{tpn}.
Besides, from the perspective of optimization, such a hierarchical scheme provides a stronger supervision, effectively avoiding the learning process from encountering issues such as gradient vanishing \cite{googlenet}, especially when a deep network is the encoder.

\subsection{Evaluation on Other Downsteam Tasks}\label{subsec:transfer}
Representations learned via supervised learning on large scale datasets such as ImageNet \cite{iamgenet} and Kinetics-400 \cite{k400} have shown to generalize well to a variety of tasks.
While previous methods for unsupervised video representation learning \
tend to study the quality of learned representations only on the action recognition task,\
it is important to include other downstream tasks for a comprehensive evaluation,\
since encoders may overfit to the action recognition benchmarks (\ie~UCF-101 \cite{ucf101} and HMDB-51 \cite{hmdb}).
Therefore, we also benchmark VTHCL on other downstream tasks, including action detection on AVA \cite{ava} and action anticipation on Epic-Kitchen \cite{epic-kitchen}.

\begin{table}[t]\centering
    \captionsetup[subfloat]{captionskip=2pt}
    \caption{
        \textbf{Representation Transfer.} Results on action detection and anticipation are reported\\
    }
    \vspace{-1em}
    \subfloat[\textbf{Action Detection on AVA.} Mean average precision (mAP) is reported\label{table:ablation:ava}]
        {
            \setlength{\tabcolsep}{1.75mm}{
        \begin{tabular}{lcccc}
            \toprule
                    &   Random  & ImageNet & Kinetics & Ours            \\
            \midrule
            R18           &  11.1     & 13.4     & 16.6     & 13.9     \\
            R50           &  7.9      & 16.8     & 21.4     & 15.0      \\
            \bottomrule
        \end{tabular}}
        }\hspace{0.5mm}
    \subfloat[\textbf{Action Anticipation on Epic-Kitchen.} Top-1 accuracy of Noun/Verb prediction is reported\label{table:ablation:epic}]
        {
            \setlength{\tabcolsep}{1.0mm}{
        \begin{tabular}{lcccc}
            \toprule
                   &   Random  & ImageNet & Kinetics & Ours            \\
            \midrule
            R18          &   8.9/26.3 & 13.5/28.0 & 14.2/28.8  & 11.2/27.0 \\
            R50          &   8.2/26.3 & 15.7/27.8 & 15.8/30.2  & 11.9/27.6    \\
            \bottomrule
        \end{tabular}}
        }
    \label{table:transfer}
    \vspace{-1em}
\end{table}

\subsubsection{Action Detection on AVA}\label{subsubsec:ava}
\noindent \textbf{Dataset.} AVA \cite{ava} provides a benchmark for spatial-temporal localization of actions.
Different from the traditional video detection (\eg ImageNet VID dataset) whose labels are categories of given bounding boxes, annotations of AVA are provided for one frame per second and describe the action over time.
AVA \cite{ava} contains around 235 training and 64 validation videos and 80 ‘atomic’ actions.

\noindent \textbf{Setup.} We follow the standard setting as in \cite{slowfast,wu2019long} for training and validation \ie we conduct the same pre-processing for region proposals.
The slow encoder $g_s$ is employed as the backbone network with the number of 8 frames as input.
Besides, the spatial stride of $res_5$ is set to 1 with the dilation of 2 to increase the spatial size of the output feature.
The region-of-interest (RoI) features are computed by 3D RoIAlign \cite{maskrcnn} and then fed into the per-class, sigmoid-based classifier for prediction.
The slight difference of training protocol is that we train our model for 24 epochs and the learning rate is decayed by a factor of 10 at 16, 22 epochs which is the standard $2\times$ scheduler of object detection.
Note that BatchNorm (BN) layers \cite{bn} are not frozen.
SGD is adopted as our optimizer with the initial learning rate of 0.1 and weight decay of $1e^{-7}$.

\noindent \textbf{Results.}
Table.\ref{table:ablation:ava} provides the mean Average Precision (mAP) of several common initialization.
Similar observation appears that with the proper initialization (\eg ImageNet, Kinetics and Ours), overfitting is slightly prevented such that 3D-ResNet50 can make the best of its increased capacity to achieve a better performance than 3D-ResNet18.
It is worth noting that our method equipped with the same backbone (13.9 mAP) can beat 3D-ResNet18 trained via supervised learning on ImageNet (13.4 mAP).
However, in action detection task, there exists a clear gap between video representations learned by self-supervised and supervised frameworks, although self-supervised approaches have obtained higher and higher results on action recognition.
It is thus beneficial and necessary to include additional downstream tasks for evaluating self-supervised video representation learning.

\subsubsection{Action Anticipation on Epic-Kitchen}\label{subsubsec:epic}
\noindent \textbf{Dataset.}
Epic-Kitchen \cite{epic-kitchen} provides a large-scale cooking dataset, which is recorded by 32 subjects in 32 kitchens.
Besides, it contains 125 verb and 352 noun categories.
Following \cite{epic-kitchen}, we randomly select 232 videos (23439 segments) for training and 40 videos (4979
segments) for validation.
Action anticipation requires to forecast the category of a future action before it happens, given a video clip as the observation.
Following the original baseline of Epic-Kitchen \cite{epic-kitchen}, we refer to $\tau_a$ as the anticipation time, and $\tau_o$ as the observation time.
In our experiments, both $\tau_a$ and $\tau_o$ are set to 1 second.

\noindent \textbf{Setup.}
In order to validate the learned representations themselves, we introduce no reasoning modules as in \cite{ke2019time, miech2019leveraging}.
Similar to \cite{epic-kitchen}, we apply a shared MLP after the backbone network and then design two separable classification heads for noun and verb predictions.
The slow encoder $g_s$ is employed as the backbone network with the number of 8 frames as input.
Our models are trained for 80 epochs with an initial learning rate of 0.1 (which is divided by 10 at 40 and 60 epoch respectively).

\noindent \textbf{Results.}
Top-1 accuracy of noun/verb prediction obtained by various models are presented in Table \ref{table:ablation:epic}.
Although our method can obtain the consistent improvements over the randomly initialized baseline, the gap between results of models learned with self-supervised and supervised schemes indicate the discriminative quality of learned representations can be further improved.

\begin{figure*}[t]
	\centering
	\includegraphics[width=1.0\textwidth]{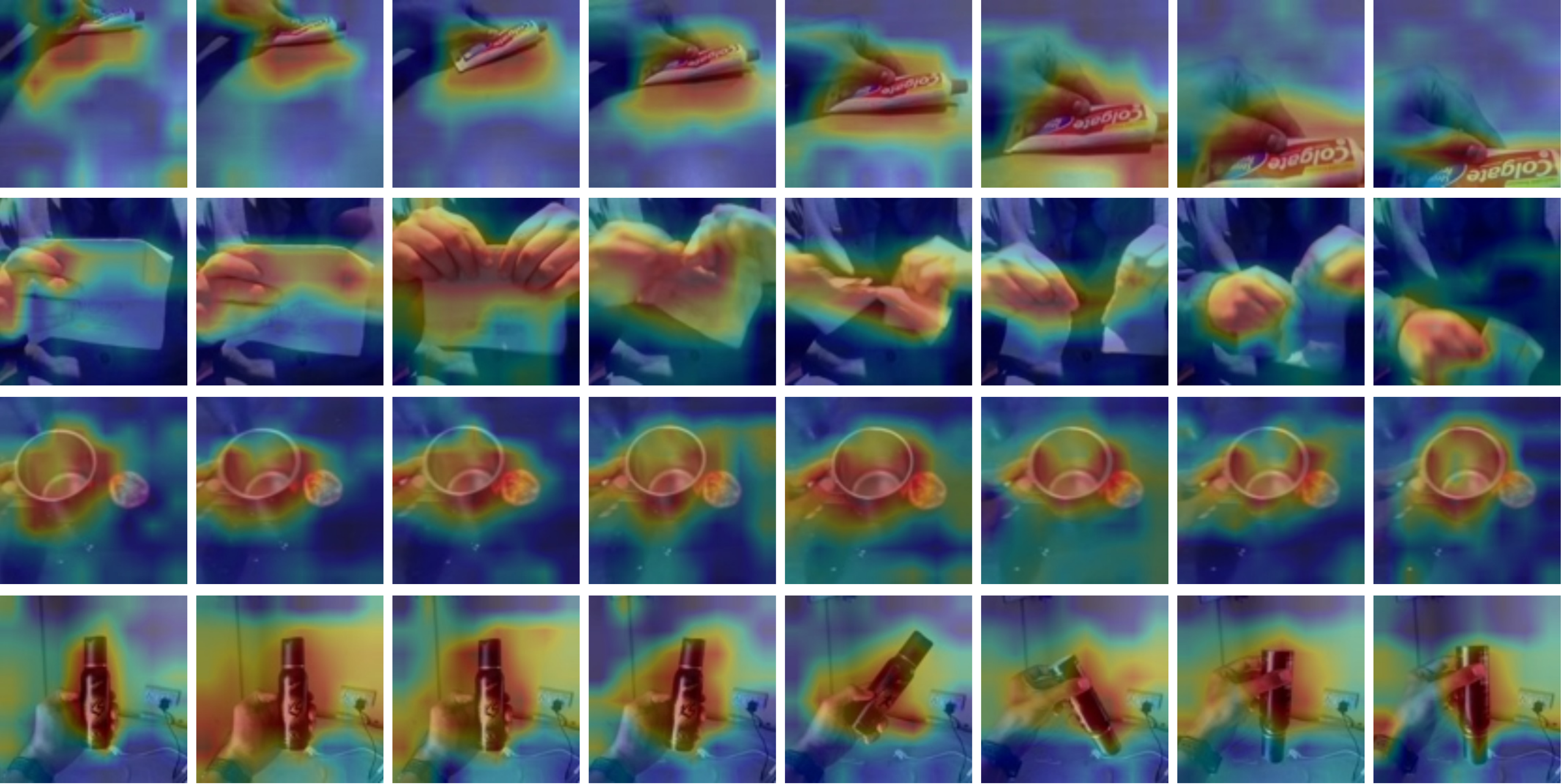}
	\captionsetup{font=small}
	\caption{
       \textbf{Examples of ICMs.}
       Without any annotations, ICM suggests that encoders try to spatially and temporally localize the core objects (\ie moving and salient objects for dynamic and static scenes respectively), when minimizing the contrastive loss
	}
    \label{fig:interpretation}
  \end{figure*}

\subsection{Discussion}\label{subsubsec:discussion}
\noindent \textbf{Heavy Backbones.}
Intuitively, heavy backbones are supposed to perform better than the lighter ones due to their increased capacity.
However, our results on action recognition, detection and anticipation reveal that heavy backbones are likely to overfit when they are not well initialized.
Therefore, when evaluating various methods of video representation learning, we should be more careful about whether they introduce consistent improvements on heavy backbones.

\noindent \textbf{Thorough Evaluation.}
From our results, we argue that we need a more thorough evaluation for learned video representations across architectures, benchmarks and downstream tasks
to study their consistency and generalization ability.
The reasons are two-fold.
a) Models with large capacities tend to overfit on UCF-101 \cite{ucf101} and HMDB-51 \cite{hmdb} due to their limited scale and diversity,\
so that augmentation and regularization sometimes can be more important than representations themselves.
Addditionally, evaluating representations for action recognition should not be the only goal.
Our study on diverse tasks shows that there remain gaps between video representations learned by self-supervised and supervised learning schemes, especially on action detection and action anticipation.
The learned representation should facilitate as many downstream tasks as possible.

\subsection{Qualitative Interpretation}\label{subsec:interpretation}
In order to investigate the shared information between slow and fast videos captured by contrast learning,
we conduct a qualitative evaluation via Instance Correspondence Map (ICM) introduced in Sec.\ref{sec:icm}.
Particularly, we train our slow and fast encoders via single-level contrastive learning on Something-Something V1 dataset \cite{sthv1} for better visualization effect.
One fully-convolutional layer is used in the learnable mapping $\phi$. 
Other settings keep the same.

Figure \ref{fig:interpretation} shows several examples of instance correspondence maps.
Although ICMs are obtained without accessing any annotations, they share large similarities semantically.
Specifically, it can be observed that our learned encoders try to localize discriminative regions spatially and temporally to distinguish instances.
In terms of instances where objects are basically static, ICM could also localize the salient objects (see the last two rows), which makes sense since the shared information between slow and fast videos are closely related to the objects.
And for those videos containing large motions, ICMs appear to capture the moving objects (\eg toothpaste and hands in the first two rows),
as motion semantics contribute more to instance classfication in these cases.
Such interpretation suggests that to some extend, semantics could emerge automatically from the proposed methods.
%

   \section{Conclusion}\label{sec:conclusion}
In this work, we leverage videos of the same instance but with varying visual tempos to learn video representations in a self-supervised learning way, \
where we adopt the contrastive learning framework and extend it to a hierarchical contrastive learning.
On a variety of downstream tasks including action recognition, detection and anticipation, we demonstrate the effectiveness of our proposed framework,
which obtains competitive results on action recognition, outperforming previous approaches by a clear margin.
Moreover, our experiments further suggest that when learning the general visual representations of videos, we should evaluate more thoroughly the learned features under different network architectures, benchmarks, and tasks.
Finally, we visualize the learned representations through the instance correspondence map to show that contrast learning on visual tempo captures the informative objects wihtout explicit supervision.

\section*{Acknowledgments}
We thank Zhirong Wu and Yonglong Tian for their public implementation of previous works.

{\small
\bibliographystyle{ieee_fullname}
\bibliography{references}
}

\end{document}